\documentclass{ifacconf}

\usepackage{bmpsize}
\usepackage[dvipdfmx]{graphicx}      
\usepackage{natbib}        
\usepackage{color}
\usepackage[utf8]{inputenc}
\usepackage{amsmath}       
\usepackage{amssymb}
\usepackage[hyphens]{url}
\usepackage[export]{adjustbox}

\newcommand{\fr}[1]{\mathsf{#1}}

\usepackage[textsize=footnotesize,textwidth=1.5cm]{todonotes}	

\newcommand{\and}[1]{\textcolor{blue}{#1}}

\definecolor{mypink1}{RGB}{255, 153, 255}

\usepackage[normalem]{ulem}

\begin{document}
\begin{frontmatter}

\title{Autonomous Landing of a \\ Multirotor Micro Air Vehicle \\
on a High Velocity Ground Vehicle\thanksref{footnoteinfo}} 

 \thanks[footnoteinfo]{This work was partially supported by CFI JELF Award 32848
 and a hardware donation from DJI.}

\author[First]{Alexandre Borowczyk} 
\author[First]{Duc-Tien Nguyen} 
\author[First]{André Phu-Van Nguyen}
\author[First]{Dang Quang Nguyen}
\author[First]{David Saussié}
\author[First]{Jerome Le Ny}

\address[First]{Mobile Robotics and Autonomous Systems Laboratory, Polytechnique Montreal and GERAD, 
   Montreal, Canada (e-mail: \{alexandre.borowczyk, duc-tien.nguyen, 
   andre-phu-van.nguyen, dang-quang.nguyen, d.saussie, jerome.le-ny\}@polymtl.ca).}

\begin{abstract}                
While autonomous multirotor micro aerial vehicles (MAVs) are uniquely well suited 
for certain types of missions benefiting from stationary flight capabilities,
their more widespread usage still faces many hurdles, due in particular to 
their limited range and the difficulty of fully automating the deployment and retrieval.
In this paper we address these issues by solving the problem of the automated
landing of a quadcopter 
on a ground vehicle moving at relatively high speed. 
We present our system architecture, including the structure 
of our Kalman filter for the estimation of the relative position and velocity 
between the quadcopter and the landing pad, as well as our controller design for 
the full rendezvous and landing maneuvers. 
The system is experimentally validated by successfully landing in multiple trials
a commercial quadcopter on the roof of a car moving at speeds of up to 50 km/h. 
\end{abstract}

\begin{keyword}
Kalman filters, Autonomous vehicles, Mobile robots, Guidance systems, Computer vision, 
Aerospace control
\end{keyword}

\end{frontmatter}

\section{Introduction}

The ability of multirotor micro aerial vehicles (MAVs) to perform stationary hover flight
makes them particularly interesting for a wide variety of applications, e.g., site surveillance,
parcel delivery, or search and rescue operations.
At the same time however, they are challenging to use on their own because of their 
relatively short battery life and short range.
Deploying and recovering MAVs from mobile Ground Vehicles (GVs) could alleviate this issue
and allow more efficient deployment and recovery in the field. For example, delivery trucks, 
public buses or marine carriers could be used to transport MAVs between locations of interest and allow 
them to recharge periodically \citep{Garone:JGCD14:UXV-GTSP, Mathew:TASE15:heterogeneousDelivery}.
For search and rescue operations, the synergy between ground and air vehicles could help save 
precious mission time and would pave the way for the efficient deployment of large fleets 
of autonomous MAVs.


The idea of better integrating GVs and MAVs has indeed already
attracted the attention of multiple car and MAV manufacturers 
\citep{mercedes2016:online, dji2016:online}. 
Research groups have previously considered the problem of landing a MAV 
on a mobile platform, but most of the existing work  
is concerned with landing on a marine platform or precision landing 
on a static or slowly moving ground target.
\cite{lange2009} provide an early example, where a custom visual marker made 
of concentric rings was created to allow for relative pose estimation,
and control was performed using optical flow and velocity commands.
More recently, \cite{yang2015} used the ArUco library from \cite{Aruco2014} 
as a visual fiducial and IMU measurements fused in a 
Square Root Unscented Kalman Filter for relative pose estimation. 
The system however still relies on optical flow for accurate velocity estimation.
This becomes problematic as soon as the MAV aligns itself with a moving ground platform, at which
point the optical flow camera suddenly measures the velocity of the platform relative 
to the MAV instead of the velocity of the MAV in the ground frame.

\cite{muskardin2016} developed a system to land a fixed wing MAV on top of a moving GV. 
However, their approach requires that the GV cooperates with the MAV during 
the landing maneuver and makes use of expensive RTK-GPS units. \cite{kim2014} show that it is possible 
to land on a moving target using simple color blob detection and a non-linear Kalman filter, but test their solution only for speeds of less than 1 m/s.
Most notably, \cite{ling2014} shows that it is possible to use low cost sensors combined 
with an AprilTag fiducial marker \citep{olson2011} to land on a small ground robot. He further 
demonstrates different methods to help accelerating the AprilTag detection.
He notes in particular that as a quadcopter pitches forward to follow the ground platform, 
the bottom facing camera frequently loses track of the visual target, 
which stresses the importance of a model-based estimator such as a Kalman filter
to compensate.

The references above address the terminal landing phase of the MAV on a moving
platform, but a complete system must also include a strategy to guide the MAV
towards the GV during its approach phase.
Proportional Navigation (PN) \citep{Kabamba14:book:GNC} is most commonly known as a 
guidance law for ballistic missiles, but can also been used for UAV guidance.
\cite{holt2010} develop a form of PN specifically for road following by a fixed-wing vehicle
and show that it is suitable for use with visual feedback coming from a gimbaled camera.
\cite{gautam2015} compare pure pursuit, line-of-sight and PN guidance laws to 
show that out of the three, PN is the most efficient in terms of the total required 
acceleration and the time required to reach the target. 
On the other hand, within close range of the target PN becomes inefficient. 
To alleviate this problem, \cite{tan2014tracking} propose a switching strategy 
to move from PN to a PD controller. Finally, to maximize the likelihood of visual 
target acquisition for a smooth transition from PN to PD, it is possible to follow 
the strategy of \cite{lin2014} to point a gimbaled camera towards a target. 



\emph{Contributions and organization of the paper.}
%
We describe a complete system allowing a multirotor MAV 
to land autonomously on a moving ground platform at relatively
high speed, using only 
commercially available and relatively low-cost sensors. 
The system architecture is described in Section \ref{section: system}. 
Our algorithms combine a Kalman filter for relative position and 
velocity estimation, described in Section \ref{section: EKF}, with 
a PN-based guidance law for the approach phase and a PID controller 
for the terminal phase. 
Both controllers are implemented using only acceleration and attitude controls, 
as discussed in Section \ref{control}.
Our design was
tested both in simulations and through extensive experiments 
with a commercially available MAV, as Section \ref{section: experiments}
illustrates. 
To the best of our knowledge, we demonstrate experimentally automatic landing 
of a multirotor MAV on a moving GV traveling at the highest 
speed to date, with successful tests carried up to a speed of 50 km/h.





\section{System Architecture}
\label{section: system}

In this section we describe the basic elements of our system architecture, 
both for the GV and the MAV. Specific details for the hardware
used in our experiments are given in Section \ref{section: experiments}.

\begin{figure}[!th]
\begin{center}
\includegraphics[bb=0.000000 0.000000 639.917980 283.967991,width=0.5\textwidth]{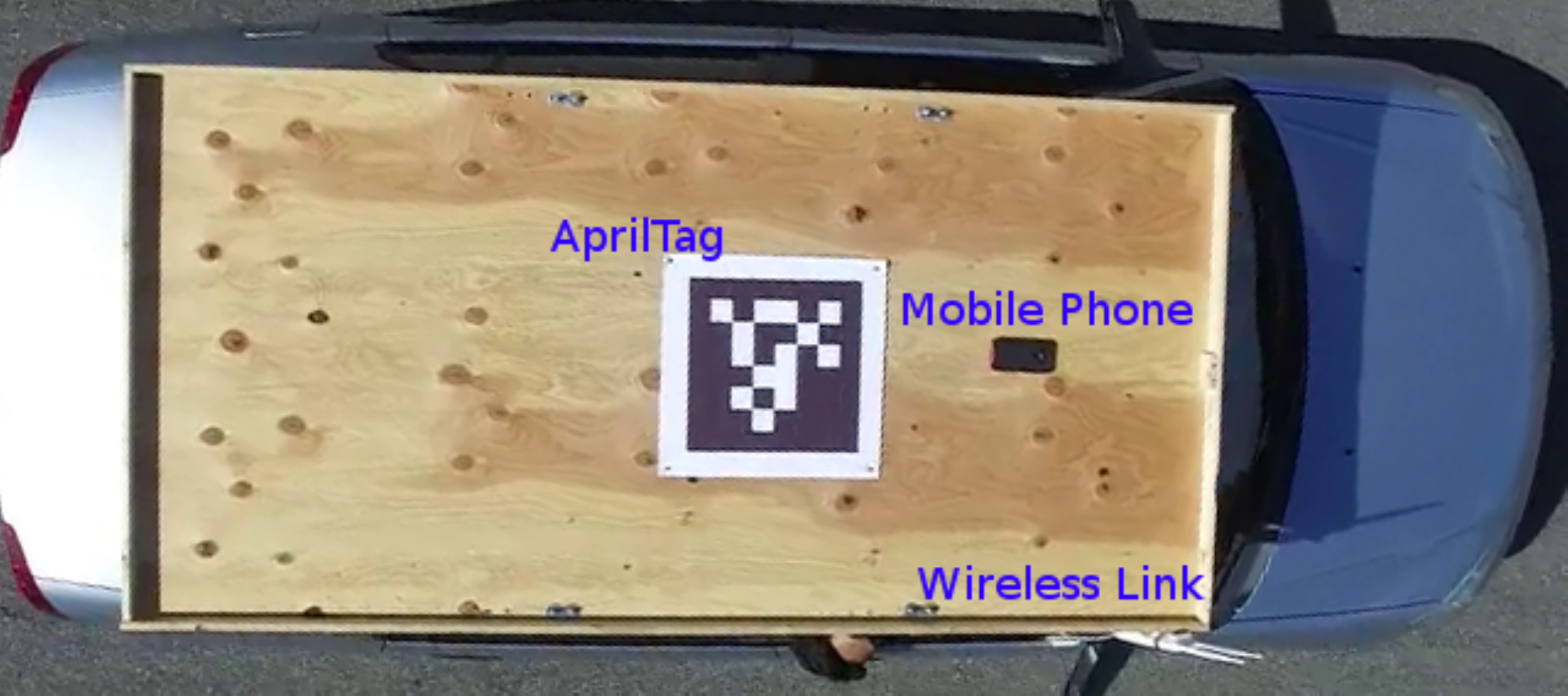}   
\caption{Experimental setup showing the required equipment on the car. In practice the mobile phone could also be held inside the car as long as GPS reception is still good.} 
\label{fig:landing_pad}
\end{center}
\end{figure}

The GV is equipped with a landing pad, on which we place
a $30 \times 30\, \mathrm{cm}$ visual fiducial named AprilTag designed by \cite{olson2011}, 
see Fig. \ref{fig:landing_pad}. 
This allows us to visually measure the 6 Degrees of Freedom (DOF) pose 
of the landing pad using onboard cameras.
In addition, we use position and acceleration measurements for the GV.
In practice, low quality sensors are enough for this purpose, and in our experiments
we simply place a mobile phone on the landing pad, which can transmit its GPS
data to the MAV at $1\, \mathrm{Hz}$ and its Inertial Measurement Unit (IMU) data at $25$ Hz at most. 
We can also integrate the rough heading and velocity estimates typically 
returned by basic GPS units, based simply on successive position measurements.

The MAV is equipped with an Inertial Navigation System (INS), an orientable $3$-axis 
gimbaled camera (with separate IMU) for target tracking purposes, as well as 
a camera with a wide angle lens pointing down, which allows us to keep track of the 
AprilTag even at close range during the very last instants of the landing maneuver. 
The approach phase can also benefit from having an additional velocity sensor on board. 
Many commercial MAVs are equipped with velocity sensors relying on optical flow methods, 
which visually estimate velocity by computing the movement of features in successive images, 
see, e.g., \citep{zhou2015}.

\begin{figure}[htp]
\begin{center}
	\includegraphics[bb=6 6 360 379,width=\linewidth]{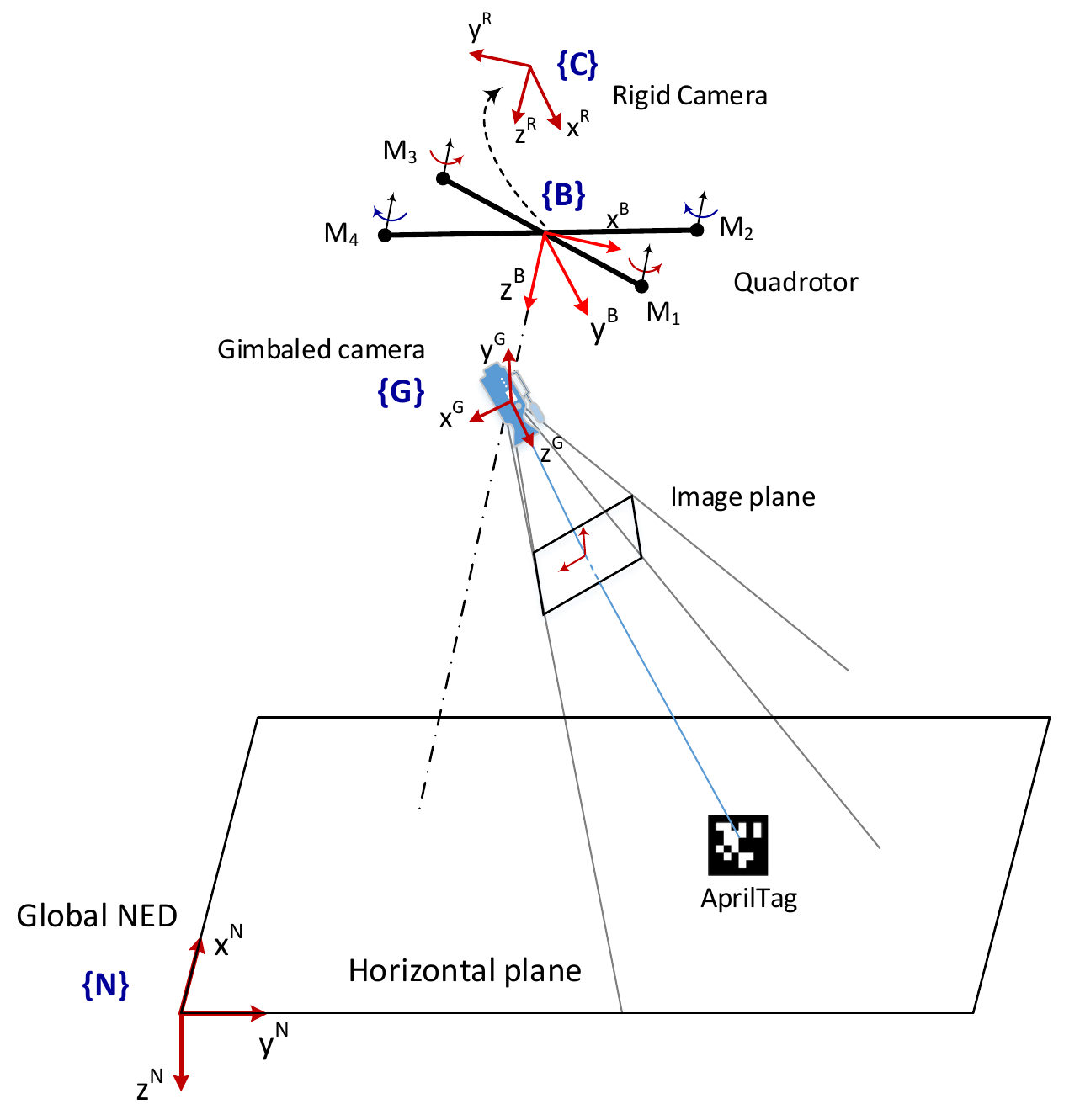} 
  	\caption{Frames of reference used} 
	\label{fig:coordinate}
\end{center}
\end{figure}

Four main coordinate frames are defined and illustrated in Fig. \ref{fig:coordinate}. 
The global North-East-Down (NED) frame, denoted $\{\fr{N}\}$, is located at the first point detected 
by the MAV. The MAV body frame $\{\fr{B}\}$ is chosen according to the cross ``$\times$'' configuration, 
i.e., its forward $x$-axis points between two of the arms and its $y$-axis points to the right.
The gimbaled camera frame $\{\fr{G}\}$ is attached to the lens center of the moving camera. Its forward 
$z$-axis is perpendicular to the image plane and its $x$-axis points to the right of the gimbal frame. 
Finally, the bottom facing rigid camera frame $\{\fr{C}\}$ is obtained from the MAV body frame by 
a $90^\circ$ rotation around the $z^{\fr{B}}$ axis and its origin is located at the optical center of camera. 

\section{Kalman filter}
\label{section: EKF}

\begin{figure}[htp]
\begin{center}
	\includegraphics[bb=5.450765 7.594031 358.638387 167.401261, width=\linewidth]{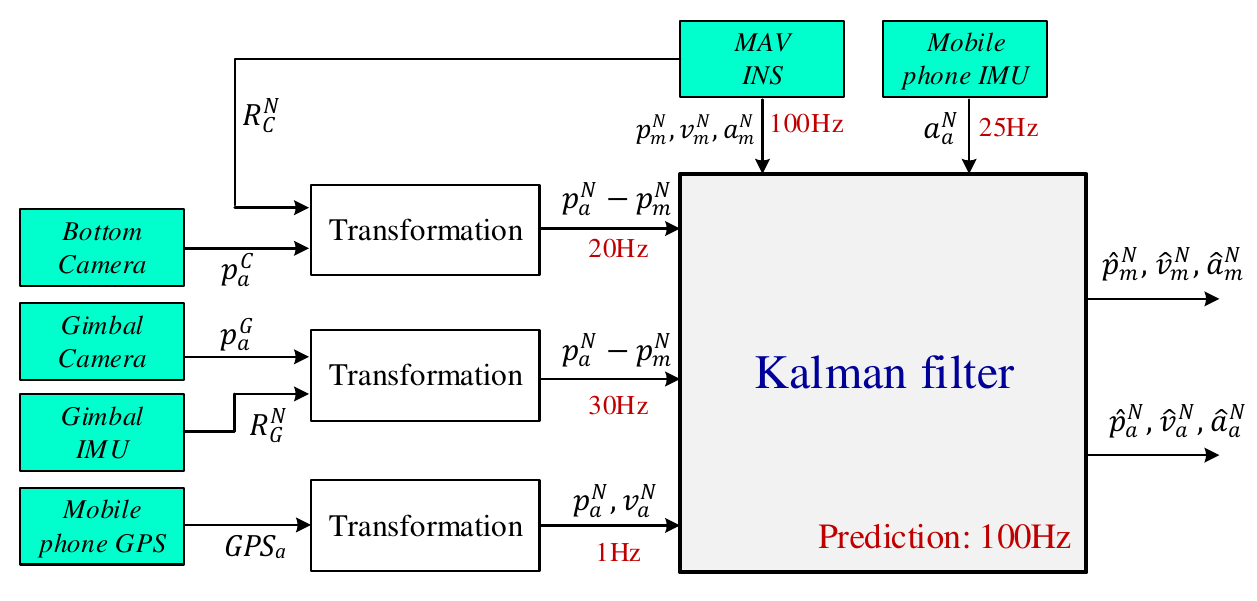} 
  	\caption{Kalman filter architecture} 
	\label{fig:architecture}
\end{center}
\end{figure}

Estimation of the relative position, velocity and acceleration between the MAV and the landing pad,
as required by our guidance and control system, 
is performed by a Kalman filter running at 100 Hz. 
The architecture of this filter is shown in Fig. \ref{fig:architecture} and 
described in the following paragraphs.

\subsection{Process model}\label{ssec:process}
In order to land on a moving target, the system estimates the three dimensional position $p(t)$, 
linear velocity $v(t)$ and acceleration $a(t)$ of the MAV and of the GV.
Thus, the state variables can be written as 
\begin{equation} \label{eq:model}
\mathbf{x} = [\mathbf{x}_m^\top \;\ \mathbf{x}_a^\top]^\top \in \mathbb{R}^{18} 
\end{equation}
where
$\mathbf{x}_m = \left[p_m^\top \;\ v_m^\top \;\ a_m^\top \right]^\top$ and 
$\mathbf{x}_a = \left[p_a^\top \;\ v_a^\top \;\ a_a^\top \right]^\top$
are respectively the state vectors for the MAV and AprilTag, 
expressed in the NED frame. The superscript $\top$ denotes the matrix transpose operation.

We use a simple second-order kinematic model for the MAV and the GV dynamics
\begin{equation}
a(t) = \ddot{p}(t) = w(t)
\end{equation}
where $w(t)$ is a white noise process with power spectral density (PSD) $q_w$. 
The corresponding discrete time model using zero-order hold (ZOH) sampling 
is given by 
\begin{equation}
\mathbf{x}_{k+1} = \mathrm{F} \mathbf{x}_k + \mathbf{w}_k
\end{equation}
\begin{displaymath}
\mathrm{F} = \begin{bmatrix}
\mathrm{F}_m & 0\\[6pt]
0 & \mathrm{F}_a\\
\end{bmatrix}, \; 
\mathrm{F}_m = \mathrm{F}_a = \begin{bmatrix}
1 & T_s & \frac{T_{s}^{2}}{2}\\[6pt]
0 & 1 & T_s\\[6pt]
0 & 0 & 1
\end{bmatrix}
\otimes \mathrm{I_3}
\end{displaymath}
where $T_s$ is the sampling period, $\otimes$ denotes the Kronecker product and 
$\mathrm{I_3}$ is the $3\times 3$ identity matrix. 
The process noise $\mathbf{w}_k$ is assumed to be a Gaussian white noise with a covariance matrix $\mathrm{Q}$ given by
\begin{align*}
\mathrm{Q} = \begin{bmatrix}
q_{w_m} & 0 \\ 0 & q_{w_a}
\end{bmatrix} \otimes \mathrm{Q_0}, \;\;
\mathrm{Q_0} = \begin{bmatrix}
\frac{T_{s}^{5}}{20} & \frac{T_{s}^{4}}{8} & \frac{T_{s}^{3}}{6}\\[6pt]
\frac{T_{s}^{4}}{8}  & \frac{T_{s}^{3}}{3}  & \frac{T_{s}^{2}}{2} \\[6pt]
\frac{T_{s}^{3}}{6}  & \frac{T_{s}^{2}}{2}  & T_s
\end{bmatrix} 
\otimes \mathrm{I}_3
\end{align*}
\noindent where $q_{w_m}$ and $q_{w_a}$ are the PSD of the MAV and GV acceleration, 
respectively. These parameters are generally chosen \textit{a priori} as part of an 
empirical tuning process.

\subsection{Measurement Model}\label{section:measurement_model}
In general, the measurement vector at time $k$ is given by 
\begin{equation} \label{eq:measurement}
\mathbf{z}_k = \mathrm{H} \mathbf{x}_k + \mathbf{v}_k,
\end{equation}
where $\mathrm{H}$ is a known matrix, $\mathbf{v}_k$ is a Gaussian white noise 
with covariance matrices $\mathrm{V}_k$ and not correlated with the process noise $\mathbf{w}_k$.
The following subsections describe the rows of the matrix $\mathrm{H}$ for the various kinds of sensor measurements,
which we simply call $\mathrm{H}$ for simplicity of notation.

\subsubsection{MAV position, velocity and acceleration from INS}
The INS of our MAV combines IMU, GPS and visual measurements to provide 
us with position, velocity and gravity compensated acceleration data directly
expressed in the global NED frame 
\begin{equation}\label{eq:mavgps}
\mathbf{z}_{k} = \begin{bmatrix} p_{m}^\top & \ v_{m}^\top & \ a_m^\top \end{bmatrix}^\top 
, \; \; 
\mathrm{H} = \begin{bmatrix}
\mathrm{I}_{9} & 0_{9 \times 9}\\
\end{bmatrix} \\ 
\end{equation}
As mentioned previously, the velocity measurement relying on optical flow methods is not correct when the MAV flies 
above a moving platform.  
Therefore, we increase the standard deviation of the velocity noise from $0.1$ m/s in the approach phase 
to $10$ m/s in the landing phase. 


\subsubsection{Target's GPS measurements} 
The GPS unit of the mobile phone on the GV provides position, speed and heading
measurements. This information is sent to the MAV on-board computer (OBC) via a 
wireless link, which gives access to the landing pad's position in the global NED frame  $\{\fr N\}$ 
\begin{align*}
&x_a^{\fr N} \approx (la - la_0)R_{E} \; , \; \;
y_a^{\fr N} \approx (lo - lo_0)\cos(la)R_{E} \\
&z_a^{\fr N} = al_0 - al,
\end{align*}
where $R_{E}=6378137\, \mathrm{m}$ is the Earth radius, $la$, $lo$, $al$ are the latitude, 
longitude (both in radians) and altitude of the landing pad respectively. 
The subscripts $_0$ corresponds to the starting point. The above equations are valid 
when the current position is not too far from the starting point and under a spherical
Earth assumption, but more precise transformations could be used \citep{Groves:Book13:nav}. 

The GPS heading $\psi_a$ and speed $U_a$ are also used to calculate 
the current velocity in the global NED frame 
\begin{equation*}
\dot{x}^{\fr N}_a =U_a\cos({\psi_a}), \;\; 
\dot{y}^{\fr N}_a =U_a\sin({\psi_a}).
\end{equation*}

So the measurement model (\ref{eq:measurement}) is expressed as follows
\begin{align*}
\mathbf{z}_{k} = 
\begin{bmatrix}
x_a^{\fr N} & y_a^{\fr N} & z_a^{\fr N} & \dot{x}_a^{\fr N} & \dot{y}_{a}^{\fr N}
\end{bmatrix}^ \top
, \; \; 
\mathrm{H} = \begin{bmatrix}
 0_{5\times 9}  & \mathrm{I}_{5} & 0_{5\times 4} 
\end{bmatrix}
\end{align*}
However, because the GPS heading measurements have poor accuracy at low speed, 
we discard them if $U_a < 2.5\,\mathrm{m/s}$, in which case
\begin{align*}
\mathbf{z}_{k} = \begin{bmatrix} x_a^{\fr N} & y_a^{\fr N} & z_a^{\fr N} \end{bmatrix} ^\top 
, \; \; 
\mathrm{H} = \begin{bmatrix}
 0_{5\times 9}  & \mathrm{I}_{3} & 0_{5\times 6} 
\end{bmatrix}
\end{align*}


The measurement noise covariance matrix $\mathrm{V}_k$ is provided by the GPS device itself. 

Our GPS receiver is a low-cost device with output rate of about 1 Hz. 
This source of data is only used to approach the target but is insufficient for landing on the moving GV. 
For the landing phase, it is necessary to use the AprilTag detection with the gimbaled and
bottom facing cameras.

\subsubsection{Gimbaled Camera measurements} 

This camera provides measurements of the relative position between the MAV and the landing 
pad with centimeter accuracy, at range up to $5$ m. 
This information is converted into the global NED frame $\{\fr N\}$ by 
\begin{equation*}
p_{m}^{\fr N} - p_{a}^{\fr N} = \mathrm{R}^{\fr N}_{\fr G}  p_{m/a}^{\fr G}
\end{equation*}
where $\mathrm{R}^{\fr N}_{\fr G}$ is the rotation matrix from $\{\fr N\}$ to $\{\fr G\}$ 
returned by the gimbal IMU. 
 Therefore, the observation model (\ref{eq:measurement}) corresponds to
\begin{gather}
\mathbf{z}_{k} =
\begin{bmatrix}
x_{m}^{\fr N} - x_{a}^{\fr N}\ & y_{m}^{\fr N} - y_{a}^{\fr N}\ &
z_{m}^{\fr N} - z_{a}^{\fr N}
\end{bmatrix}^\top  \nonumber \\
\mathrm{H} = \begin{bmatrix}
\mathrm{I}_{3} & 0_{3\times 6}  
& -\mathrm{I}_{3} & 0_{3\times 6} 
\end{bmatrix}. \nonumber
\label{eq:camera_measure}
\end{gather}

Here the standard deviation of the measurement noise is empirically set to $0.2$ m. 
To reduce the chances of target loss, the gimbaled camera centers the image onto the AprilTag as soon as visual detection is achieved. When the AprilTag cannot be detected, we follow  the control scheme proposed by \cite{lin2014} to point the camera towards the landing pad  using the estimated line-of-sight (LOS) information obtained from the Kalman filter.

\subsubsection{Bottom camera measurements} 
The bottom facing camera is used to assist the last moments of the landing, 
when the MAV is close to the landing pad, yet too far to cut off the motors. 
At that moment, the gimbaled camera can not perceive the whole AprilTag 
but a wide angle camera like the mvBlueFOX can still provide measurements.

This camera measures the target position in the frame $\{\fr C\}$, 
as illustrated in Fig. \ref{fig:coordinate}. Hence, the observation 
model is the same as (\ref{eq:camera_measure}), except for the transformation  
to the global NED frame 
\begin{equation*}
p_{m}^{\fr N} - p_{a}^{\fr N} = \mathrm{R}^{\fr N}_{\fr C} ~ p_{m/a}^{\fr C}  
\end{equation*}
where 
$\mathrm{R}^{\fr N}_{\fr C}$ denotes the rotation matrix from $\{\fr N\}$ to $\{\fr C\}$.


\subsubsection{Landing pad acceleration using the mobile phone's IMU}
Finally, since most mobile phones also contain an IMU, we leverage this sensor to estimate the GV's acceleration.
\begin{gather*}
\mathbf{z}_{k} = \begin{bmatrix} \ddot{x}_{a}^{\fr N} & \ddot{y}_{a}^{\fr N} & \ddot{z}_{a}^{\fr N}
\end{bmatrix}^\top
, \; \; 
\mathrm{H} = \begin{bmatrix}
0_{3\times 15} & \mathrm{I}_{3}\\
\end{bmatrix} \\ 
\mathrm{V}_k = {\rm diag}(0.6^2, \, 0.6^2, \, 0.6^2) ~ \text{m/s}^2.
\end{gather*}

The Kalman filter algorithm follows the standard two steps, 
with the prediction step running at $100$ Hz and the measurement update step 
executed as soon as new measurements become available. 
The output of this filter is the input to the guidance 
and control system described in the next section, 
which is used by the MAV to approach and land safely on the moving platform. 

\section{Guidance and Control System}
\label{control}


For GV tracking by the MAV, we use a guidance strategy switching between a Proportional Navigation (PN) law \citep{Kabamba14:book:GNC} for the approach phase and a PID for the landing phase,
which is similar in spirit to the approach in \citep{tan2014tracking}.


The approach phase is characterized by a large distance between the MAV and the GV 
and the absence of visual localization data. Hence, in this phase,
the MAV has to rely on the data transmitted by the GV's GPS and IMU. 
The goal of the controller for this phase is to follow an efficient
``pursuit'' trajectory, which is achieved here by a PN controller 
augmented with a closing velocity controller.
In contrast, the landing phase is characterized by a relatively close 
proximity between the MAV and the GV and the availability 
of visual feedback to determine the target's position. 
This phase requires a higher level of accuracy and faster response time 
from the controller, and a PID controller can be more easily tuned to
meet these requirements than a PN controller.
%
In addition, the system should transition from one controller to the other 
seamlessly, avoiding discontinuity in the commands sent to the MAV.


\emph{Proportional Navigation Guidance}
The well-known PN guidance law uses the fact that two vehicles 
are on a collision course if their LOS remains at 
a constant angle in order to steer the MAV toward the GV.
It works by keeping rotation of the velocity vector proportional 
to the rotation of the LOS vector.
Our PN controller provides an acceleration command 
that is normal to the instantaneous LOS
\begin{equation}	\label{eq: perp. PN controller}
a_{\bot} = - \lambda |\dot{u}|\frac{u}{|u|}\times \Omega,
\; \text{with } \Omega = \frac{u \times \dot{u}}{u \cdot u},
\end{equation}

where $\lambda$ is a gain parameter, 
$u = p_a^{\fr N} - p_m^{\fr N}$ and
$\dot u = v_a^{\fr N} - v_m^{\fr N}$
are obtained from the Kalman filter and
represent the LOS vector and its derivative
expressed in the NED frame, and 
$\Omega$ is the rotation vector of the LOS. 
We then supplement the PN guidance law \eqref{eq: perp. PN controller} 
with an approach velocity controller determining the acceleration $a_{\parallel}$ 
along the LOS direction, 
which in particular allows us to specify a high enough velocity required 
to properly overtake the target. 
This acceleration component is computed using the following PD structure
\begin{align*}
a_{\parallel} = K_{p_{\parallel}}u  + K_{d_{\parallel}} \dot{u}
\end{align*}
where $K_{p_{\parallel}}$ and $K_{i_{\parallel}}$ are constant gains.
The total acceleration command is obtained by combining both components
\begin{equation*}
a = a_{\bot} + a_{\parallel}.
\end{equation*}
As only the horizontal control is of interest, the acceleration along $z$-axis is disregarded. The desired acceleration then needs to be converted to attitude control inputs
that are more compatible with the MAV input format. In frame $\{\fr N\}$, the quadrotor dynamic equations of translation read as follows 
\begin{equation*}
m a_m  =
\begin{bmatrix}0 \\ 0 \\ mg\end{bmatrix} + 
\mathrm{R}^{\fr N}_{\fr B} \begin{bmatrix}0 \\ 0 \\ -T\end{bmatrix} + 
F_{D}
\end{equation*}
where $\mathrm{R}^{\fr N}_{\fr B}$ denotes the rotation matrix from $\{\fr N\}$ to $\{\fr B\}$,
\begin{equation*}
\mathrm{R}^{\fr N}_{\fr B} = 
\begin{bmatrix}
c_{\theta} c_{\psi} &
 s_{\phi}s_{\theta}c_{\psi}- c_{\phi} s_{\psi}  &
 c_{\phi}s_{\theta}c_{\psi}+ s_{\phi} s_{\psi}  \\
c_{\theta} s_{\psi} &
 s_{\phi} s_{\theta} s_{\psi} + c_{\phi} c_{\psi}  &
 c_{\phi}s_{\theta}s_{\psi} - s_{\phi} c_{\psi} \\
-s_{\theta} &
s_{\phi} c_{\theta} &
c_{\phi} c_{\theta}
\end{bmatrix},
\end{equation*}
$T$  the total thrust created by rotors, $F_{D}$ the drag force, $m$  the MAV mass and $g$ the standard gravity.
The force equations simplify as
\begin{equation*}
m\begin{bmatrix} \ddot{x}_{m} \\  \ddot{y}_{m} \\  \ddot{z}_{m}\end{bmatrix} = 
\begin{bmatrix}0 \\ 0 \\ m g\end{bmatrix} - 
\begin{bmatrix}
 c_{\phi}s_{\theta}c_{\psi}+ s_{\phi} s_{\psi} \\
c_{\phi}s_{\theta}s_{\psi} - s_{\phi} c_{\psi} \\
c_{\phi} c_{\theta}
\end{bmatrix}  T + 
\begin{bmatrix}-k_d \dot{x}_m|\dot{x}_m| \\-k_d \dot{y}_m|\dot{y}_m| \\ -k_d \dot{z}_m|\dot{z}_m|\end{bmatrix}
\end{equation*}
where the drag is roughly modeled as a force proportional to the signed quadratic velocity in each direction and $k_d$ is constant, which we estimated  by recording the terminal velocity for a range of attitude controls at level flight and performing a least squares regression on the data. For constant flight altitude, $T = { m g}/{ c_{\phi} c_{\theta}}$ and assuming $\psi = 0$, it yields
\begin{equation*}
m\begin{bmatrix}\ddot{x}_{m} \\  \ddot{y}_{m}  \end{bmatrix} =
mg
\begin{bmatrix}  - \tan {\theta} \\
   \tan {\phi}  / \cos {\theta}
\end{bmatrix}  - 
\begin{bmatrix}k_d \dot{x}_m|\dot{x}_m| \\ k_d \dot{y}_m|\dot{y}_m|\end{bmatrix} 
\end{equation*}
The following relations can then be obtained:
\begin{align*}
\theta  & = -\arctan\left( (m\ddot{x}_{m} + k_d \dot{x}_m|\dot{x}_m|)/ mg\right)\\
\phi & =  \arctan\left(   \cos {\theta} \left(m \ddot{y}_{m} + k_d \dot{y}_m|\dot{y}_m| \right)/ mg\right), 
\end{align*}
where $\theta$ and $\phi$ are the desired pitch and roll angles for specific acceleration demands.  


\emph{PID controller}
The landing phase is handled by a PID controller, with
the desired acceleration computed as 
\begin{equation*}
a = K_p u +  K_i\int{u} + K_d \dot{u},
\end{equation*}
where $K_p$, $ K_i $ and $K_d$ are constant gains.
The tuning for the PID controller was selected to provide aggressive dynamic path following, 
promoting a quick disturbance rejection. The controller was first tuned in simulation 
and then the settings were manually adjusted in flight.

\emph{Controller switching}
The controller switching scheme chosen is a simple fixed distance switching condition 
with a slight hysteresis. The switching distance selected was $6\, \mathrm{m}$ to allow for 
any perturbation due to the switching to dissipate before reaching the landing platform.

\emph{Vertical control}
The entire approach phase is done at a constant altitude, which is handled by the internal vertical position controller of the MAV. The descent is initiated once the quadrotor has stabilized over the landing platform. A constant vertical velocity command is then issued to the MAV and maintained until it reaches a height of $0.2\, \mathrm{m}$ above the landing platform at which point the motors are disarmed.

\section{Experimental Validation}
\label{section: experiments}

\subsection{System Description}

\begin{figure}[!ht]
\begin{center}
\includegraphics[bb=0.000000 0.000000 626.993981 469.997986,width=8.4cm]{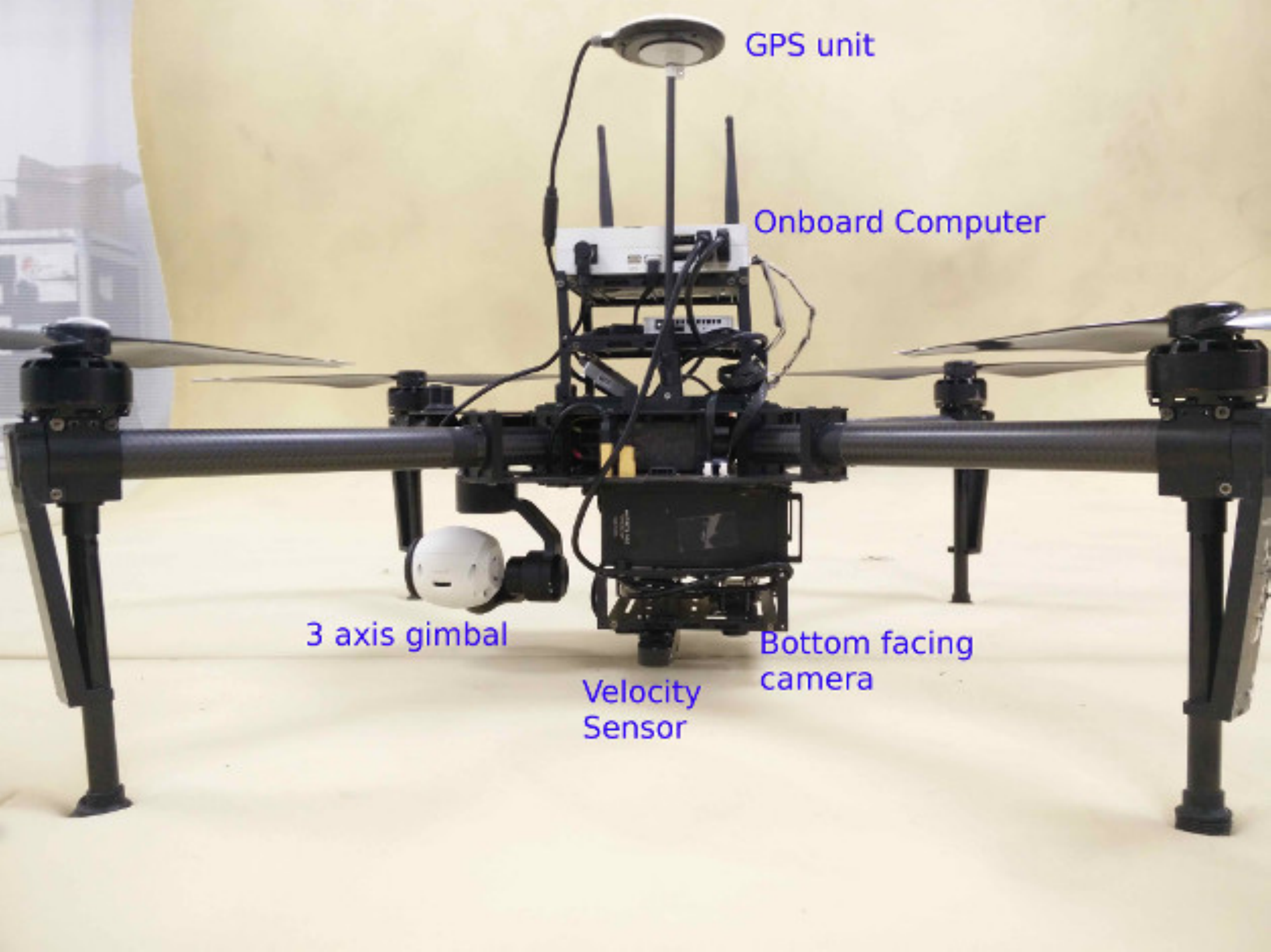}    
\caption{The M100 quadcopter. Note that all side facing cameras of the
Guidance module were removed and the down facing BlueFox camera sits behind the bottom guidance sensor.
} 
\label{fig:m100}
\end{center}
\end{figure}

We implemented our system on a commercial off-the-shelf DJI Matrice 100 (M100) quadcopter 
shown in Fig. \ref{fig:m100}.
All computations are performed on the standard OBC for this platform (DJI Manifold), 
which contains an Nvidia Tegra K1 SoC.
The 3-axis gimbaled camera is a Zenmuse X3, from which we receive 720p YUV color images 
at 30 Hz. To reduce computations, we drop the U and V channels and downsample the images 
to obtain $640\times 360$ monochrome images. 
We modified the M100 to rigidly attach a downward facing Matrix Vision mvBlueFOX camera, 
equipped with an ultra-wide angle Sunex DSL224D lens with a diagonal field of view of $176$ degrees.
The M100 is also equipped with the DJI Guidance module, an array of up to five stereo cameras, 
which are meant to help developers create mapping and obstacle avoidance solutions for
robotics applications. 
This module seamlessly integrates with the INS to provide us with position, velocity and acceleration measurements for the M100 using a fusion of on-board sensors described in \cite{zhou2015}. This information is used in our Kalman filter in equation \eqref{eq:mavgps}.
%
%

Our algorithms were implemented in C++ using ROS \citep{ROS:ICRA09}. 
Using an open source implementation of the AprilTag library based on OpenCV 
and with accelerations provided by OpenCV4Tegra, we can run the tag detection 
at a full 30 fps using the X3 camera and 20 fps on the BlueFOX camera. 

As mentioned in Section \ref{control}, we implemented our control system using pure
attitude control in the $xy$ axes and velocity control in the $z$ axis. The reason for
this is that the internal velocity estimator of the M100 fuses optical flow measurements 
from the Guidance system. These measurements become extremely inaccurate once the quadcopter 
flies over the landing platform. Although optical flow could be used to measure the relative 
velocity of the car, it would be difficult to pinpoint the moment where flow measurements 
transition from being with respect to the ground to being with respect to the moving car.

\subsection{Experimental Results}

\begin{figure}[htp]
\begin{center}    
    \includegraphics[width=\textwidth/2]{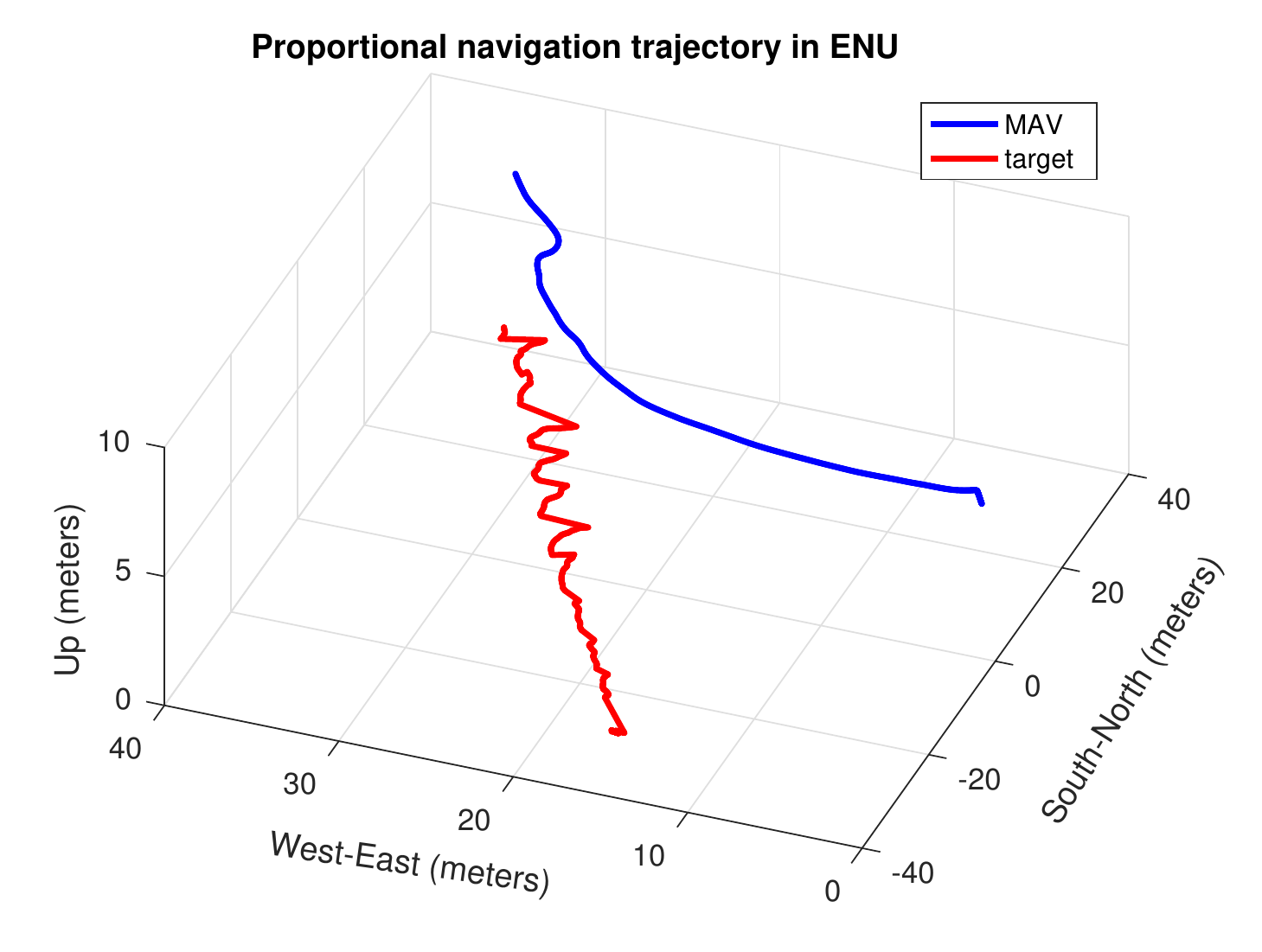}
  	\caption{The PN controller efficiently catches up with the target at long range even when the only source of information is the mobile phone's GPS and IMU.}
	\label{fig:3Dpn}
\end{center}
\end{figure}

A first experiment was done using only data from the mobile phone (no visual feedback) to prove the validity of our Proportional Navigation controller for the long range approach. Figure \ref{fig:3Dpn} shows the output of our EKF estimating the MAV's and the target's positions. With the target and the MAV starting at about 30 meters from each other, we see the MAV fly a smooth rendezvous trajectory eventually ending on top of the target.

The close range system was experimentally validated with the GV moving at speeds as low as a human jogging 
and on a private race track at 30, 40 and 50 km/h with successful landings in each case. 
Videos of our experiments can be found at 
\url{https://youtu.be/ILQqD2xQ4tg}.

\begin{figure}[htp]
\begin{center}
    \includegraphics[bb=9.684000 5.238000 499.805985 291.959991,width=\textwidth/2]{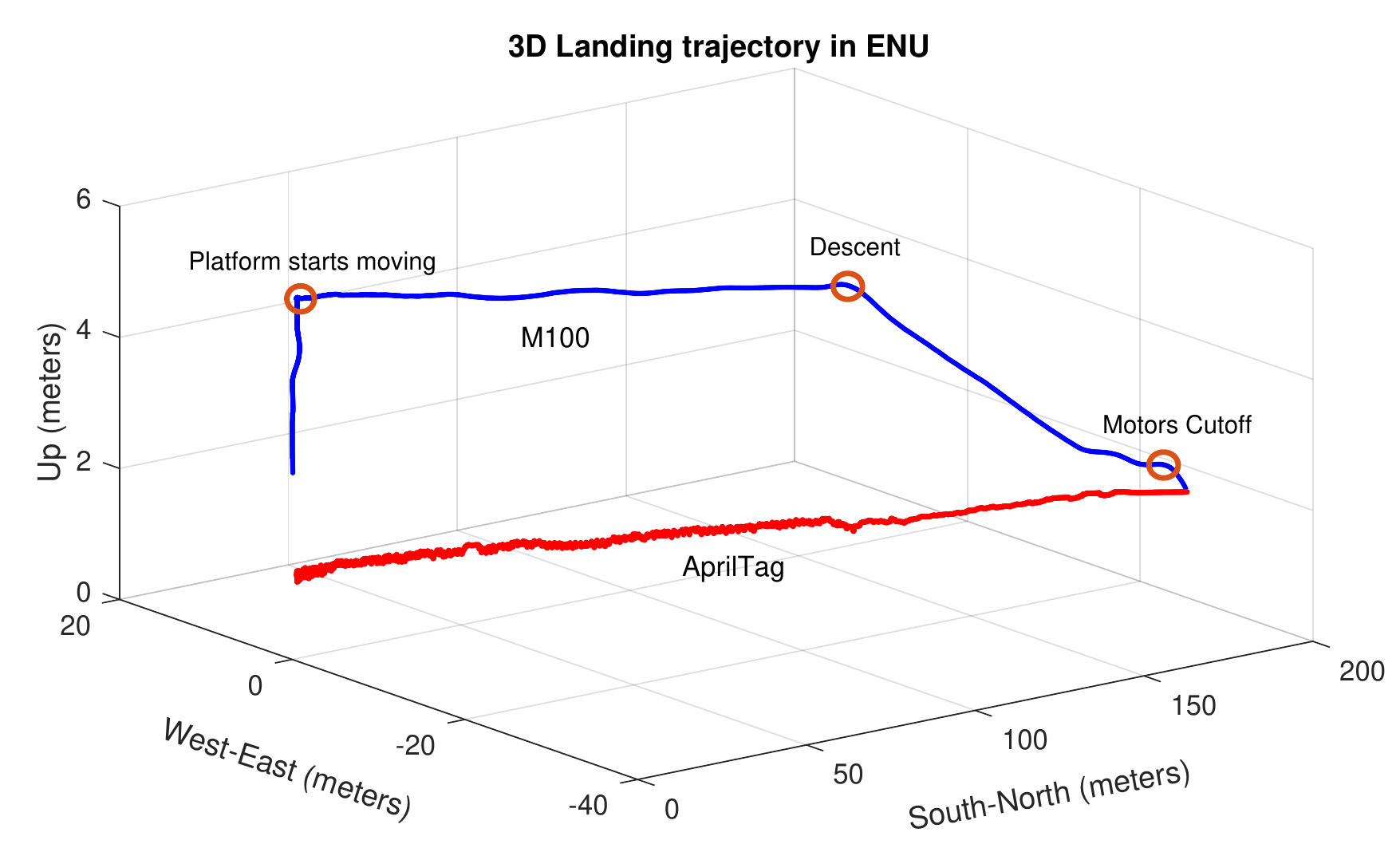}
  	\caption{Landing trajectory at 50km/h using the PID controller.}
	\label{fig:3Dposition}
\end{center}
\end{figure}

\begin{figure}[htp]
\begin{center}
    \includegraphics[bb=2.772000 1.152000 593.459982 233.477993,scale=0.36]{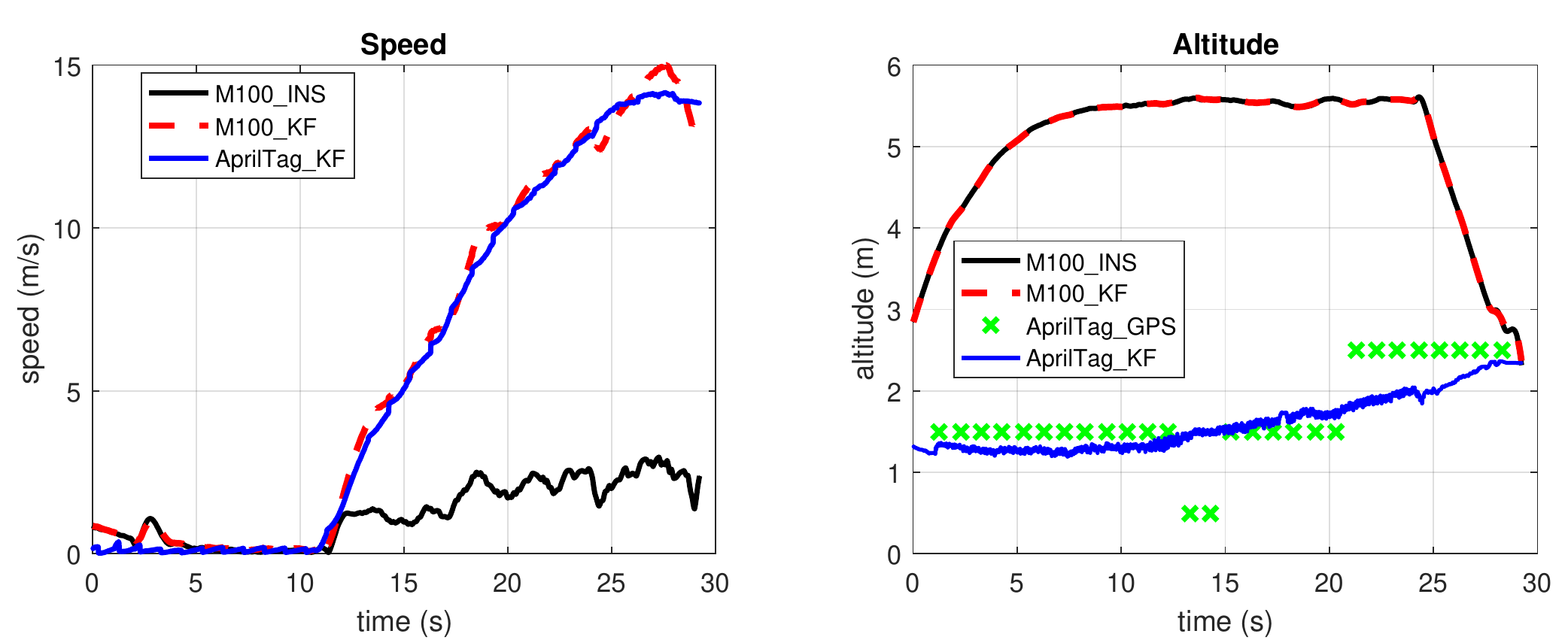}
  	\caption{Motions of M100 and AprilTag.}
	\label{fig:motions}
\end{center}
\end{figure}

Figure \ref{fig:3Dposition} shows the landing sequence where the
quadrotor takes off, tracks and lands on the landing pad. The curves gain in altitude
as the trajectory progresses because of the elevation profile of the race track. The effect
is seen more clearly in Fig. \ref{fig:motions}, where we can also see the filtered 
AprilTag altitude rise, thanks to visual data and the M100's internal altitude estimator, 
even before the phone's GPS data indicates a change in altitude. 

\begin{figure}[htp]
\begin{center}
    \includegraphics[bb=9.054000 4.878000 475.631985 300.185991, scale=0.465]{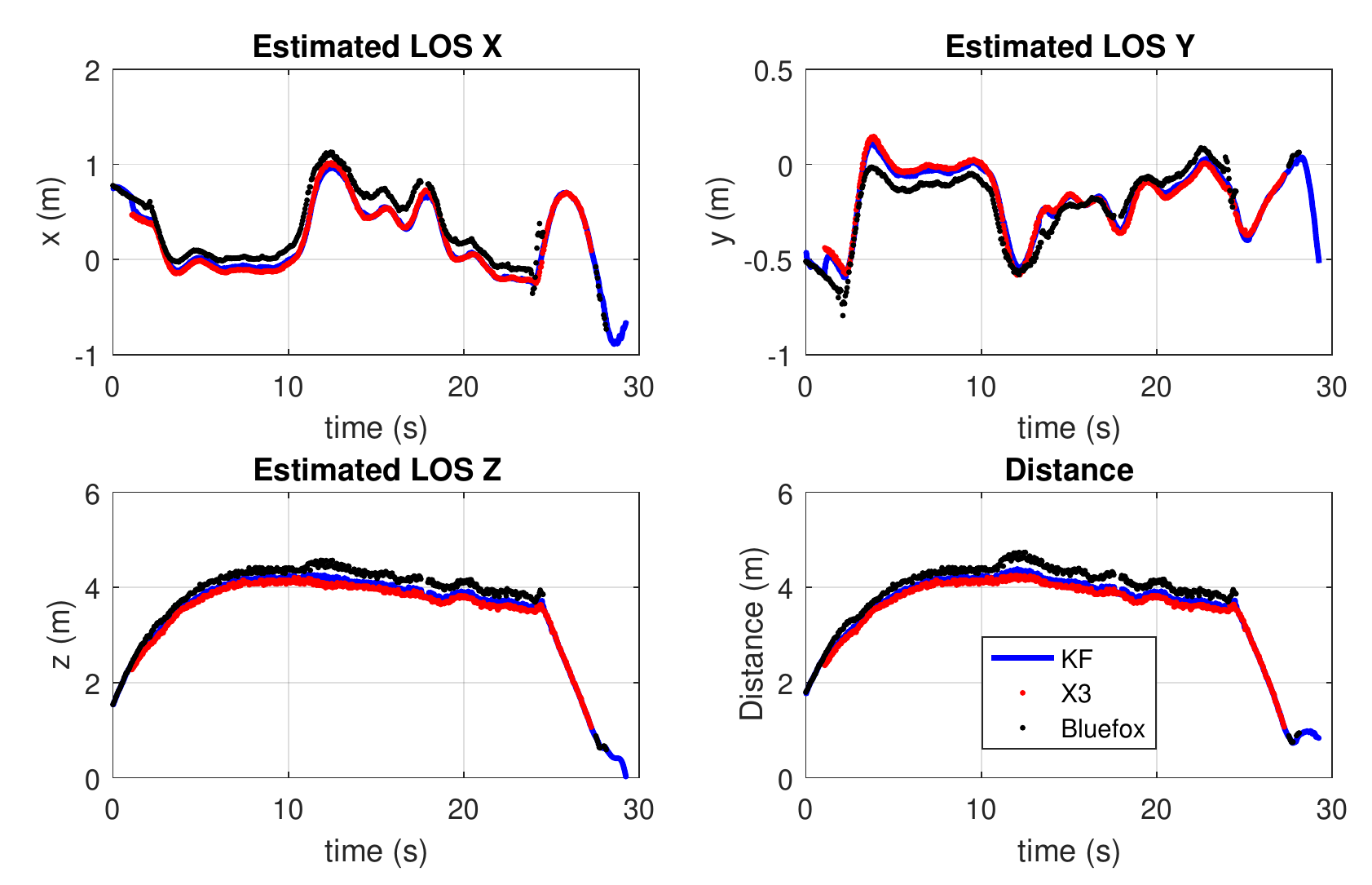}
  	\caption{LOS and distance (coordinates in NED frame).}
	\label{fig:LOS}
\end{center}
\end{figure}

\begin{figure}[htp]
\begin{center}
    \includegraphics[width=\textwidth/2]{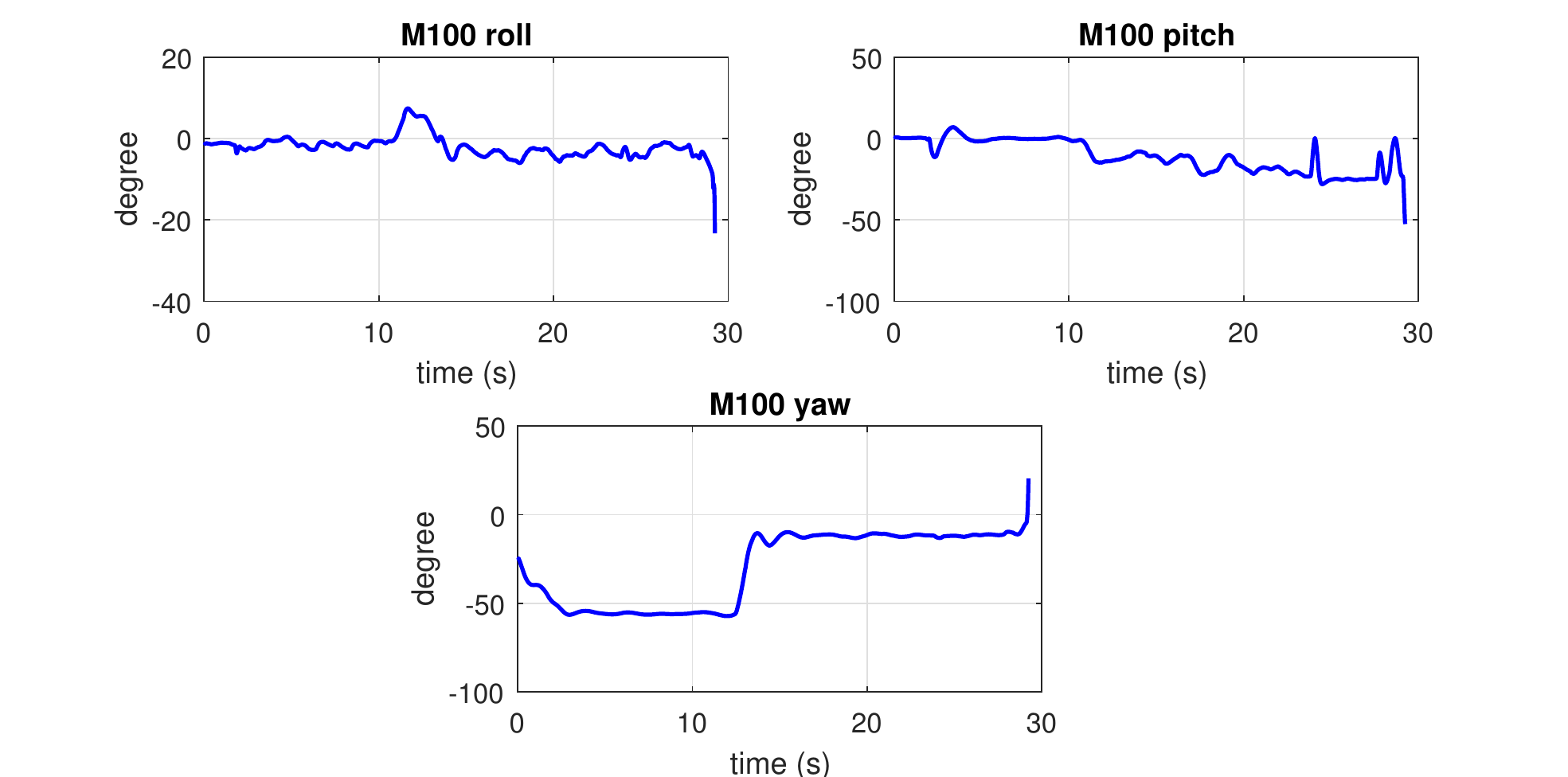}
  	\caption{M100 attitude.}
	\label{fig:attitude}
\end{center}
\end{figure}

Furthermore, we can see in Fig. \ref{fig:motions} how the M100 closely matches
the velocity of the AprilTag to perform the landing maneuver. The two peaks at the
24 and 27 second marks are strongly correlated to the visual loss 
of the tag by the BlueFOX camera which we can observe in Fig. \ref{fig:LOS}.
The descent starts at the 24 second mark slightly before the car hits its
designated velocity of 14 m/s or 50.4 km/h. We can also see in Fig. \ref{fig:motions} how the M100's velocity estimation from the INS becomes incorrect when it is on top of the car. Which explains why we decided to dynamically increase the standard deviation of the measurement as described in Section \ref{section:measurement_model}. Figure \ref{fig:attitude} shows the 
quadcopter's attitude during the flight. Notice how the roll is stable close to $0^\circ$
 while the pitch stays between $-10^\circ$ and $-25^\circ$ for consistent forward
flight. Finally, the yaw changes drastically after 10 seconds at the moment 
the car starts moving.

\section{Conclusion}

The problem of the automatic landing of a MAV on a moving vehicle was solved with experimental tests going up to 50 km/h. A Proportional Navigation controller was used for the long range approach, which subsequently transitioned into a PID controller at close range. A Kalman filter was used to estimate the position of the MAV relative to the landing pad by fusing together measurements from the MAV's onboard INS, a visual fiducial marker and a mobile phone. Furthermore, it was shown that this system can be implemented using only commercial off-the-shelf components.

Future work may include using a better multiscale visual fiducial on the landing pad to allow visual target tracking at longer and closer ranges using a single camera or simplifying the system by removing the requirement for a mobile phone. Performance improvements could also be achieved by adding a model of the ground vehicle's turbulence or adding wind estimation in the control system.





\bibliography{ifacconf}             

\begin{thebibliography}{19}
\providecommand{\natexlab}[1]{#1}
\providecommand{\url}[1]{\texttt{#1}}
\providecommand{\urlprefix}{URL }
\expandafter\ifx\csname urlstyle\endcsname\relax
  \providecommand{\doi}[1]{doi:\discretionary{}{}{}#1}\else
  \providecommand{\doi}{doi:\discretionary{}{}{}\begingroup
  \urlstyle{rm}\Url}\fi

\bibitem[{Garone et~al.(2014)Garone, Determe, and
  Naldi}]{Garone:JGCD14:UXV-GTSP}
Garone, E., Determe, J.F., and Naldi, R. (2014).
\newblock Generalized traveling salesman problem for carrier-vehicle systems.
\newblock \emph{AIAA Journal of Guidance, Control and Dynamics}, 37(3),
  766--774.

\bibitem[{Garrido-Jurado et~al.(2014)Garrido-Jurado, noz Salinas,
  Madrid-Cuevas, and Mar\'in-Jim\'enez}]{Aruco2014}
Garrido-Jurado, S., noz Salinas, R.M., Madrid-Cuevas, F., and
  Mar\'in-Jim\'enez, M. (2014).
\newblock Automatic generation and detection of highly reliable fiducial
  markers under occlusion.
\newblock \emph{Pattern Recognition}, 47(6), 2280 -- 2292.

\bibitem[{Gautam et~al.(2015)Gautam, Sujit, and Saripalli}]{gautam2015}
Gautam, A., Sujit, P.B., and Saripalli, S. (2015).
\newblock Application of guidance laws to quadrotor landing.
\newblock In \emph{Unmanned Aircraft Systems (ICUAS), 2015 International
  Conference on}, 372--379.

\bibitem[{Groves(2013)}]{Groves:Book13:nav}
Groves, P.D. (2013).
\newblock \emph{Principles of {GNSS}, inertial, and multisensor integrated
  navigation systems}.
\newblock Artech House, 2nd edition.

\bibitem[{Holt and Beard(2010)}]{holt2010}
Holt, R. and Beard, R. (2010).
\newblock Vision-based road-following using proportional navigation.
\newblock \emph{Journal of Intelligent and Robotic Systems}, 57(1-4), 193 --
  216.

\bibitem[{Kabamba and Girard(2014)}]{Kabamba14:book:GNC}
Kabamba, P.T. and Girard, A.R. (2014).
\newblock \emph{Fundamentals of Aerospace Navigation and Guidance}.
\newblock Cambridge University Press.

\bibitem[{Kim et~al.(2014)Kim, Jung, Lee, and Shim}]{kim2014}
Kim, J., Jung, Y., Lee, D., and Shim, D.H. (2014).
\newblock Outdoor autonomous landing on a moving platform for quadrotors using
  an omnidirectional camera.
\newblock In \emph{Unmanned Aircraft Systems (ICUAS), 2014 International
  Conference on}, 1243--1252.

\bibitem[{Kolodny(2016)}]{mercedes2016:online}
Kolodny, L. (2016).
\newblock Mercedes-benz and {M}atternet unveil vans that launch delivery
  drones.
\newblock \url{http://tcrn.ch/2c48his}.
\newblock (Accessed on 09/09/2016).

\bibitem[{Lange et~al.(2009)Lange, Sunderhauf, and Protzel}]{lange2009}
Lange, S., Sunderhauf, N., and Protzel, P. (2009).
\newblock A vision based onboard approach for landing and position control of
  an autonomous multirotor uav in gps-denied environments.
\newblock In \emph{Advanced Robotics, 2009. ICAR 2009. International Conference
  on}, 1--6.

\bibitem[{Lardinois(2016)}]{dji2016:online}
Lardinois, F. (2016).
\newblock Ford and {DJI} launch \$100,000 developer challenge to improve
  drone-to-vehicle communications.
\newblock \url{tcrn.ch/1O7uOFF }.
\newblock (Accessed on 09/28/2016).

\bibitem[{Lin and Yang(2014)}]{lin2014}
Lin, C.E. and Yang, S.K. (2014).
\newblock Camera gimbal tracking from uav flight control.
\newblock In \emph{Automatic Control Conference (CACS), 2014 CACS
  International}, 319--322.

\bibitem[{Ling(2014)}]{ling2014}
Ling, K. (2014).
\newblock \emph{Precision Landing of a Quadrotor UAV on a Moving Target Using
  Low-cost Sensors}.
\newblock Master's thesis, University of Waterloo.

\bibitem[{Mathew et~al.(2015)Mathew, Smith, and
  Waslander}]{Mathew:TASE15:heterogeneousDelivery}
Mathew, N., Smith, S.L., and Waslander, S.L. (2015).
\newblock Planning paths for package delivery in heterogeneous multirobot
  teams.
\newblock \emph{IEEE Transactions on Automation Science and Engineering},
  12(4), 1298--1308.

\bibitem[{Muskardin et~al.(2016)Muskardin, Balmer, Wlach, Kondak, Laiacker, and
  Ollero}]{muskardin2016}
Muskardin, T., Balmer, G., Wlach, S., Kondak, K., Laiacker, M., and Ollero, A.
  (2016).
\newblock Landing of a fixed-wing uav on a mobile ground vehicle.
\newblock In \emph{2016 IEEE International Conference on Robotics and
  Automation (ICRA)}, 1237--1242.

\bibitem[{Olson(2011)}]{olson2011}
Olson, E. (2011).
\newblock Apriltag: A robust and flexible visual fiducial system.
\newblock In \emph{Robotics and Automation (ICRA), 2011 IEEE International
  Conference on}, 3400--3407.

\bibitem[{Quigley et~al.(2009)Quigley, Conley, Gerkey, Faust, Foote, Leibs,
  Wheeler, and Ng}]{ROS:ICRA09}
Quigley, M., Conley, K., Gerkey, B.P., Faust, J., Foote, T., Leibs, J.,
  Wheeler, R., and Ng, A.Y. (2009).
\newblock {ROS}: an open-source robot operating system.
\newblock In \emph{{ICRA} Workshop on Open Source Software}.

\bibitem[{Tan and Kumar(2014)}]{tan2014tracking}
Tan, R. and Kumar, M. (2014).
\newblock Tracking of ground mobile targets by quadrotor unmanned aerial
  vehicles.
\newblock \emph{Unmanned Systems}, 2(02), 157--173.

\bibitem[{Yang et~al.(2015)Yang, Ying, Lu, and Li}]{yang2015}
Yang, S., Ying, J., Lu, Y., and Li, Z. (2015).
\newblock Precise quadrotor autonomous landing with {SRUKF} vision perception.
\newblock In \emph{2015 IEEE International Conference on Robotics and
  Automation (ICRA)}, 2196--2201.

\bibitem[{Zhou et~al.(2015)Zhou, Fang, Tang, Zhang, Wang, and Yang}]{zhou2015}
Zhou, G., Fang, L., Tang, K., Zhang, H., Wang, K., and Yang, K. (2015).
\newblock Guidance: A visual sensing platform for robotic applications.
\newblock In \emph{2015 IEEE Conference on Computer Vision and Pattern
  Recognition Workshops (CVPRW)}, 9--14.

\end{thebibliography}
  
\end{document}